\documentclass[11pt, letterpaper]{article}
\usepackage[utf8]{inputenc}
\usepackage[margin=1in]{geometry}
\usepackage{graphicx}
\usepackage{authblk}
\graphicspath{ {./} }

\makeatletter
  \def\title@font{\Large\bfseries}
  \let\ltx@maketitle\@maketitle
  \def\@maketitle{\bgroup%
    \let\ltx@title\@title%
    \def\@title{\resizebox{\textwidth}{!}{%
      \mbox{\title@font\ltx@title}%
    }}%
    \ltx@maketitle%
  \egroup}
\makeatother
 
\title{Skin Lesion Segmentation Using Atrous Convolution via DeepLab v3}
\author{Yujie Wang$^{\dag}$ - Troy High School, MI, wangyuji431@gmail.com \and Simon Sun$^{\dag}$ - Harvard College, ssun@college.harvard.edu \and Jahow Yu - Troy High School, MI, jahowyu@gmail.com \and Dr. Limin Yu - Beaumont Health, limin.yu@beaumont.edu \and $^{\dag}$ Authors contributed equally}
\date{July 2018}
 
\begin{document}
 
\maketitle
 
\begin{abstract}
	As melanoma diagnoses increase across the US, automated efforts to identify malignant lesions become increasingly of interest to the research community. Segmentation of dermoscopic images is the first step in this process, thus accuracy is crucial. Although techniques utilizing convolutional neural networks have been used in the past for lesion segmentation, we present a solution employing the recently published DeepLab 3, an atrous convolution method for image segmentation. Although the results produced by this run are not ideal, with a mean Jaccard index of 0.498, we believe that with further adjustments and modifications to the compatibility with the DeepLab code and with training on more powerful processing units, this method may achieve better results in future trials.
\end{abstract}

\section{Introduction}
	More than 5 million cases of skin cancer are diagnosed in the United States each year, surpassing diagnoses of all other cancers combined. As new diagnoses have increased by 53\% in the last decade, it is imperative that medical professionals are able to properly recognize and identify malignant lesions. Dermatologists frequently rely on techniques such as the ABCDE rule (checking if the lesion is Asymmetric, has an irregular Border, has a uniform Color, whether the Diameter is greater than 6 mm, and whether the lesion Evolves and changes over time) to accomplish this task. However, with the advances of image processing and machine learning technologies, one growing interest among the research community is automatic lesion segmentation, feature classification, and disease classification. 
	
The International Skin Imaging Collaboration (ISIC), sponsored by the International Society for Digital Imaging of the Skin, has hosted its annual challenge since 2016, divided among those three sub-tasks. As a part of the 2018 challenge, we present a lesion segmentation technique using DeepLab 3, a semantic image segmentation model utilizing atrous convolution in convolutional neural networks (CNNs), implemented in Google’s Tensorflow framework. As segmentation is the first step in diagnosis, it is especially important that accuracy is maintained-- many structural factors of diagnosis, such as asymmetry and border shape, are extracted from this phase. 

Neural networks offer themselves as a natural solution to the image segmentation problem. A key fixture among machine learning applications, neural networks are composed of a number of hidden layers, an input layer, and an output layer, with each layer composed of an adjustable amount of nodes. In supervised learning, which our method uses, the network is fed training data, the results of which are compared to a given ground truth. Based on the error between the expected values of the output nodes and the returned values, the relations between each node are adjusted.This repeated process continues as the network approaches its maximum accuracy. In a sense, we can treat neural networks as a black box, only worrying about certain hyperparameters such as the number and type of layers, number of times the test data is run, batch size, and learning rate.
Convolutional neural networks (CNNs) are a special type of neural network, most often used in image processing. These networks have two specialized kinds of layers, namely convolution layers and pooling layers, as well as a fully-connected layer. Convolution layers apply a filter to the input matrix to output a feature map. These layers are then followed by any number of pooling layers, which retains the key features while reducing the dimensionality of the input. Finally, the fully-connected layer relate the extracted features to the various traits we are hoping to classify. The number, order, and combinations of all these layers are adjustable and thus affect the overall effectiveness of the CNN.

The top submission from the 2017 edition of the ISIC challenge used CNN variations to produce segmentation predictions with an average Jaccard index of 0.765. Our approach aims to accomplish the goal using the 2017 release of DeepLab 3. This uses atrous convolution, in which the filter is not applied to all adjacent pixels of an image but rather a spaced-out lattice of pixels. This allows the resulting feature map to retain spacial resolution without significant cost to computation time. DeepLab uses several of these convolutions in cascade, followed by a “Pyramid Pooling” method in which four parallel atrous convolutions are applied to the input, each with differing atrous rates. Using these strategies, DeepLab 3 was able to perform significantly better than previous DeepLab versions and had comparable performances to other methods on the PASCAL VOC 2012 dataset.

Previous lesion segmentation methods have employed CNNs, but have not utilized the architecture of DeepLab 3. For instance, the 2016 effort by Mohammad H. Jafari et al. utilized two parallel paths of calculation, one for local analysis and one for general analysis, each with two pairs of convolution and a max pooling layer. These two paths are then followed by a fully-connected layer before the output layer. This method was able to achieve 98\% accuracy on the Dermquest dataset, better than other algorithms such as Otsu.

\section{Approach}
	Our method was trained and tested on the ISIC Archive dataset, which contains quality-controlled dermoscopic images of lesions. This means that there is little if any preprocessing that needs to be done on the input images before entering into the CNN. Additionally, each training data item is provided with a ground truth image, so computing error and performing stochastic descent is made easier.
	
	In order to make the dataset compatible with the DeepLab code, the images were resized to 513x513 and then converted into TFRecord. Various options regarding network model were available, including ResNet and Xception. Additionally, various hyperparameters were adjustable, including weighting of foreground and background, learning rate, and learning rate decay. 
	
	Most of the challenge was involved in adapting the dataset to successfully train the model. As we wished to use the existing DeepLab code, our efforts went into adjusting and debugging the code to allow the model to accept the ISIC images. We received multiple errors on this front, but for our most successful run, we initialized using the weights in the checkpoint used to segment the Pascal dataset with 0.001 learning rate and foreground loss weight 100 for 100 steps. We believe that training for more steps would help us achieve better results.
	
In order to evaluate the model, we initially worked to run the built in evaluate function contained in DeepLab used to evaluate the example models on our trained dataset. However, after various errors, we decided to write our own evaluation script to calculate a Jaccard value for each image by counting the tuples of pixels of each color in the prediction and ground truth images. We ran this on 963 ISIC images to give us generalized data about the model.

	In order to finalize the prediction images to fit the ISIC challenge criteria, we wrote a script to recolor the prediction images and upscaled them back to the original size using OpenCV. This does result in more pixelated images, but we believe that using a more powerful computer may allow running the neural network on larger image sizes, which may improve accuracy.
	
\begin{center}
\begin{figure}[t]
\centering
\caption{A sampling of segmentation prediction results (left of image) compared to ground truths (right). Jaccard indexes from top to bottom: 0.943, 0.875, 0.271, 0.527}

\includegraphics[width=0.16\textwidth]{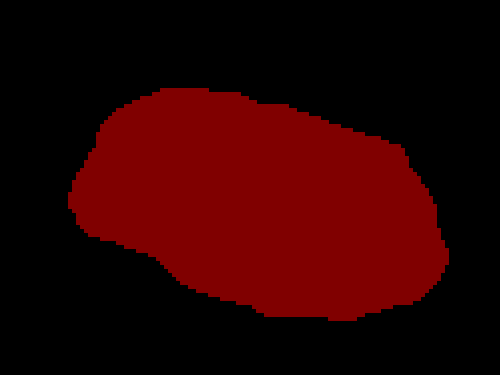}
\includegraphics[width=0.16\textwidth]{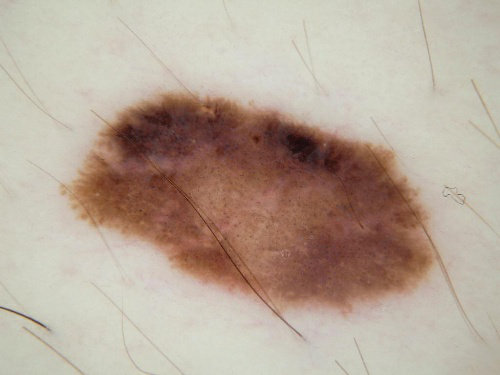}
\includegraphics[width=0.16\textwidth]{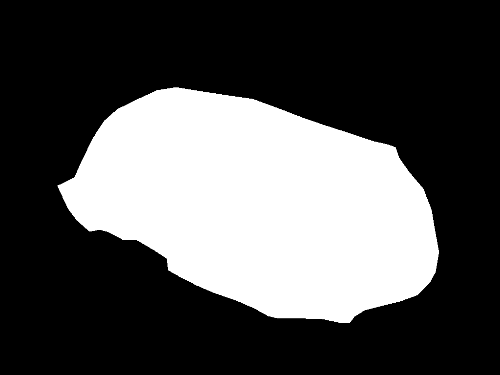}

\includegraphics[width=0.16\textwidth]{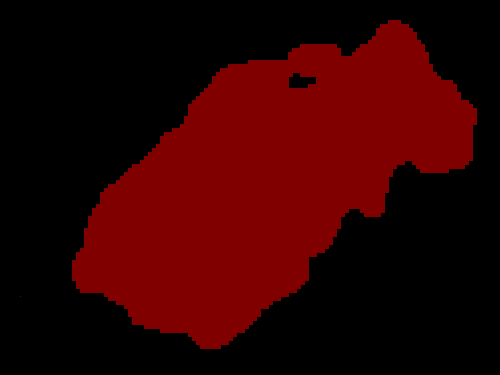}
\includegraphics[width=0.16\textwidth]{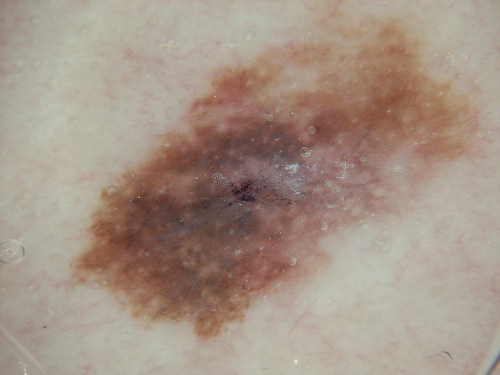}
\includegraphics[width=0.16\textwidth]{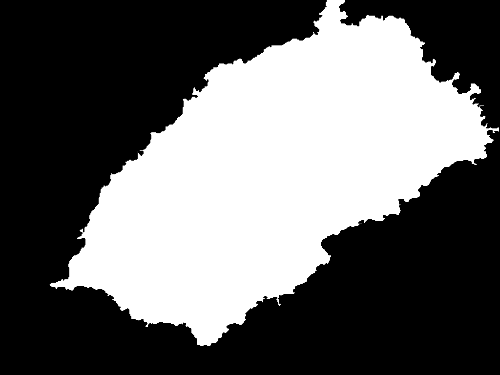}

\includegraphics[width=0.16\textwidth]{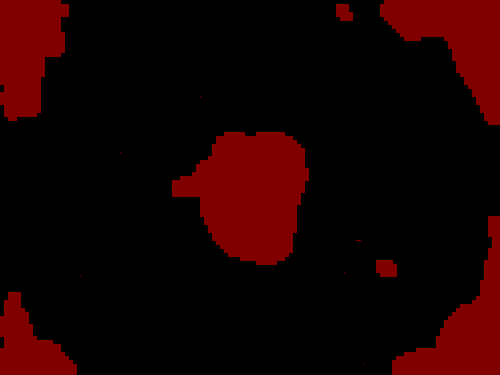}
\includegraphics[width=0.16\textwidth]{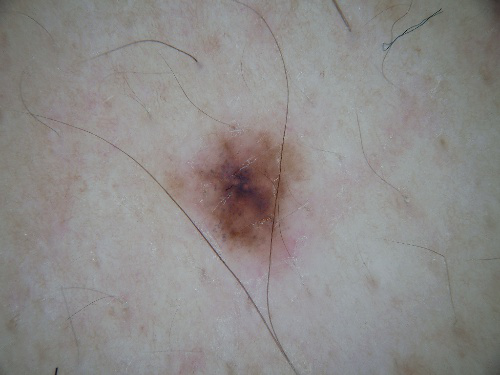}
\includegraphics[width=0.16\textwidth]{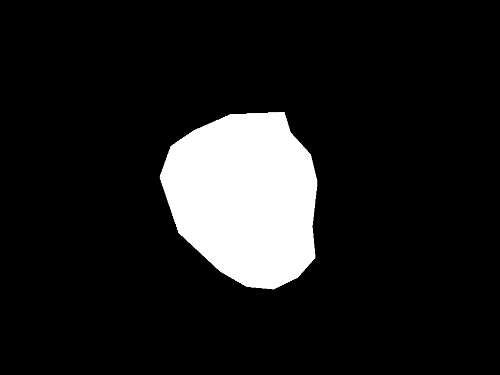}

\includegraphics[width=0.16\textwidth]{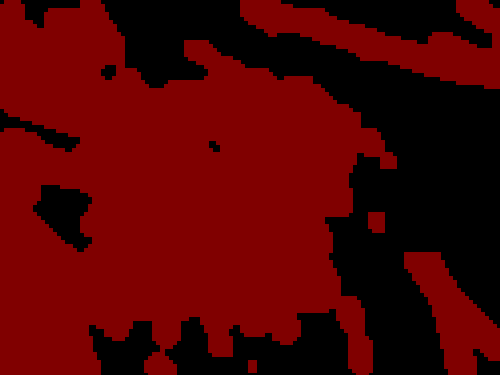}
\includegraphics[width=0.16\textwidth]{000906_image}
\includegraphics[width=0.16\textwidth]{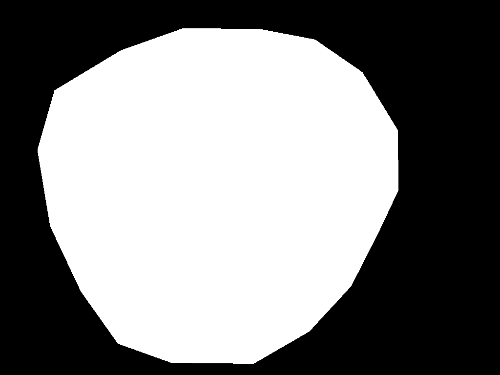}
\end{figure}
\end{center}

\section{Results}
In our most successful run, we achieved a mean Jaccard value of 0.498, median Jaccard value of 0.500, and standard deviation of 0.273. Unfortunately, ISIC considers Jaccard values under 0.650 to be unsegmented, meaning that overall, most of the images are unsuccessful. Of 963 images we evaluated on, 335 pass this threshold, giving us a 34.787\% success rate using this method.

From a subjective standpoint, we noticed that our neural network worked best on images that were clear, had even lighting, and were free of other objects such as hair or follicles. Further training may improve these results. We believe that with further compatibility with the DeepLab framework, training for more steps (say, 100,000), and more time to allow tweaking of hyperparameters, perhaps using stochastic gradient descent, that the results may improve to an  acceptable successful result. Much of the error in our result we believe to be a consequence of our implementation rather than the method itself.

\section{Conclusion}
	To segment skin lesions for the 2018 ISIC challenge, we attempted an approach that used atrous convolution in a convolutional neural network. Specifically, we adapted the DeepLab v3 code to formulate and train a model. Despite many compatibility issues, we ultimately succeeded in running the dataset through it. While the results leave much room for improvement, with mean Jaccard value of 0.498, we are optimistic that further testing can improve the segmentation success rate to a favorable level.


\begin{thebibliography}{9}
\bibitem{chen} 
Liang-Chieh Chen, George Papandreou, Florian Schroff, Hartwig Adam. 
"Rethinking Atrous Convolution for Semantic Image Segmentation", 2017; 
arXiv:1706.05587.

\bibitem{isic}
Noel C. F. Codella, David Gutman, M. Emre Celebi, Brian Helba, Michael A. Marchetti, Stephen W. Dusza, Aadi Kalloo, Konstantinos Liopyris, Nabin Mishra, Harald Kittler, Allan Halpern. “Skin Lesion Analysis Toward Melanoma Detection: A Challenge at the 2017 International Symposium on Biomedical Imaging (ISBI), Hosted by the International Skin Imaging Collaboration (ISIC)”, 2017; arXiv:1710.05006.

\bibitem{jafari}
Mohammad H. Jafari, Ebrahim Nasr-Esfahani, Nader Karimi, S.M. Reza Soroushmehr, Shadrokh Samavi, Kayvan Najarian.
“Extraction of Skin Lesions from Non-Dermoscopic Images Using Deep Learning”, 2016;
arXiv:1609.02374

\bibitem{isic2}
Tschandl, P., Rosendahl, C. \& Kittler, H. The HAM10000 dataset, a large collection of multi-source dermatoscopic images of common pigmented skin lesions. Sci. Data 5, 180161 doi:10.1038/sdata.2018.161 (2018).

\bibitem{knuthwebsite} 
The Skin Cancer Foundation,
\\\texttt{https://www.skincancer.org/skin-cancer-information/skin-cancer-facts}
\end{thebibliography}
\end{document}